\documentclass{article}

    \usepackage[preprint,nonatbib]{neurips_2024}

\usepackage[utf8]{inputenc} 
\usepackage[T1]{fontenc}    
\usepackage[hidelinks]{hyperref}       
\usepackage{url}            
\usepackage{booktabs}       
\usepackage{amsfonts}       
\usepackage{nicefrac}       
\usepackage{microtype}      
\usepackage{xcolor}         

\usepackage{mathtools}
\mathtoolsset{showonlyrefs}

\usepackage{amssymb} 
\usepackage{float} 
\usepackage{subcaption} 
\usepackage{graphicx} 
\graphicspath{{../FIGURES/}}

\usepackage{enumitem}

\usepackage{mdframed}
\definecolor{theoremcolor}{rgb}{255, 255, 255}
\mdfsetup{
    backgroundcolor=theoremcolor,
    linewidth=0.5pt,
}

\newmdtheoremenv{definition}{Definition}
\newmdtheoremenv{proposition}{Proposition}
\newmdtheoremenv{corollary}{Corollary}
\newmdtheoremenv{theorem}{Theorem}
\newmdtheoremenv{lemma}{Lemma}
\newmdtheoremenv{example}{Example}

\newcommand{\norm}[1]{\left\lVert #1\right\rVert}
\newcommand{\Indicator}[1]{\iota_{#1}}  
\def\lsc{{l.s.c.\ }}
\def\aka{{a.k.a.\ }}
\def\wrt{{w.r.t.\ }}
\def\ones{\mathbf{1}}

\def\cC{{\mathcal{C}}}
\def\cD{{\mathcal{D}}}

\def\RR{{\mathbb{R}}}
\def\RRb{{\overline{\mathbb{R}}}}
\def\e{{\mathrm{e}}}

\DeclareMathOperator*{\dom}{dom}
\DeclareMathOperator*{\argmax}{argmax}
\DeclareMathOperator*{\argmin}{argmin}

\title{Learning with Fitzpatrick Losses}

\author{%
  Seta Rakotomandimby \\
  Ecole des Ponts\\
  \texttt{seta.rakotomandimby@enpc.fr} \\
  \And
  Jean-Philippe Chancelier \\
  Ecole des Ponts\\
  \texttt{jean-philippe.chancelier@enpc.fr} \\
  \And
  Michel De Lara \\
  Ecole des Ponts\\
  \texttt{michel.delara@enpc.fr} \\
  \And
  Mathieu Blondel \\
  Google DeepMind \\
  \texttt{mblondel@google.com} \\
}

\begin{document}

\maketitle

\begin{abstract}
Fenchel-Young losses are a family of convex loss functions,
encompassing the squared, logistic and sparsemax losses, among others.
Each Fenchel-Young loss is implicitly associated with a link function, for
mapping model outputs to predictions. For instance, the logistic loss is
associated with the soft argmax link function. Can we build new loss functions
associated with the same link function as Fenchel-Young losses?
In this paper, we introduce Fitzpatrick losses, a new family of convex loss
functions based on the Fitzpatrick function. A well-known theoretical tool in
maximal monotone operator theory, the Fitzpatrick function naturally leads to a
refined Fenchel-Young inequality, making Fitzpatrick losses tighter than
Fenchel-Young losses, while maintaining the same link
function for prediction.  
As an example, we introduce the Fitzpatrick logistic loss and the
Fitzpatrick sparsemax loss, counterparts of the logistic and the sparsemax
losses. This yields two
new tighter losses associated with the soft argmax and the sparse argmax,
two of the most ubiquitous output layers used in machine learning. We study in
details the properties of Fitzpatrick losses and in particular, we show that
they can be seen as Fenchel-Young losses using a modified, target-dependent
generating function. We demonstrate the effectiveness of Fitzpatrick losses for
label proportion estimation.
\end{abstract}

\section{Introduction}

Loss functions are a cornerstone of statistics and machine learning: they
measure the difference, or ``loss,'' between a ground-truth target and a
model prediction. As such, they have attracted a wealth of research.
Proper losses (\aka proper scoring rules)
\cite{grunwald2004game,gneiting2007strictly} 
measure the discrepancy between
a target distribution and a probability forecast.
They are essentially primal-primal Bregman divergences, with both the target and
the prediction belonging to the same primal space.
They are typically explicitly composed with a link function
\cite{reid2010composite,williamson2016composite}, in order to map the model
output to a prediction. A disadvantage of this
explicit composition is that it often makes the resulting composite loss
function nonconvex. A related family of loss functions are Fenchel-Young losses
\cite{fyl_aistats,fyl_jmlr}, which encompass many commonly-used loss functions in
machine learning including the squared, logistic, sparsemax and perceptron
losses. Fenchel-Young losses can be seen as primal-dual Bregman divergences
\cite{amari2016information}, with the target belonging to the primal space and
the model output belonging to the dual space. In contrast to proper
losses, each Fenchel-Young loss is implicitly associated with a given link function, 
mapping the dual-space model output to a
primal-space prediction (for instance, the soft argmax is the link function
associated with the logistic loss). This crucial difference makes
Fenchel-Young losses always convex.
Can we build new convex losses associated with the same link function as
Fenchel-Young losses?

\begin{figure}[t]
\centering
\includegraphics[width=0.48\textwidth]{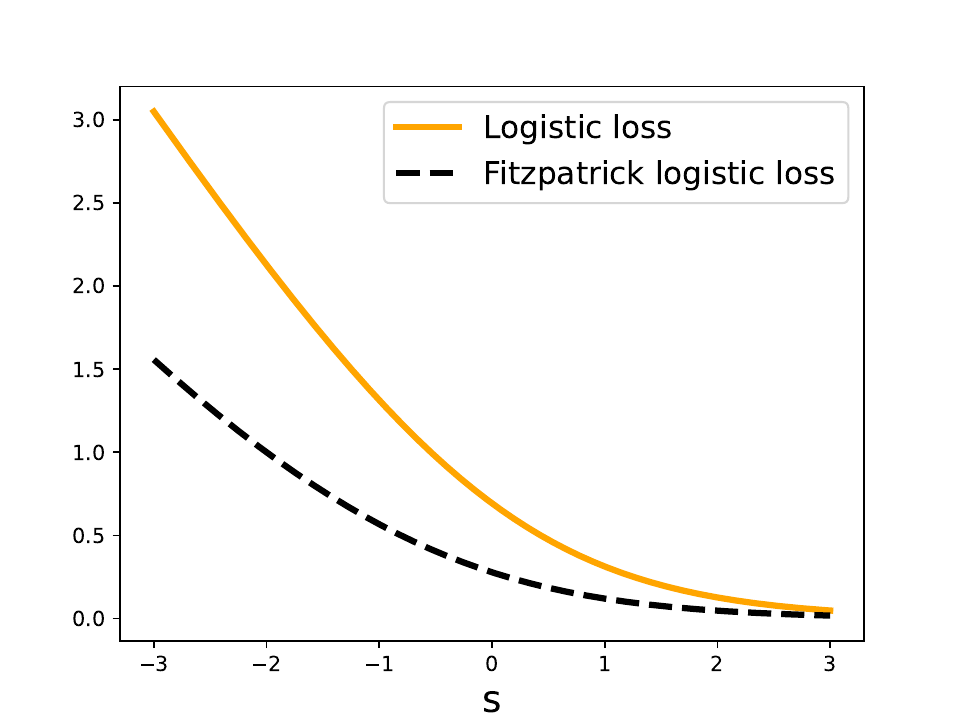}
\includegraphics[width=0.48\textwidth]{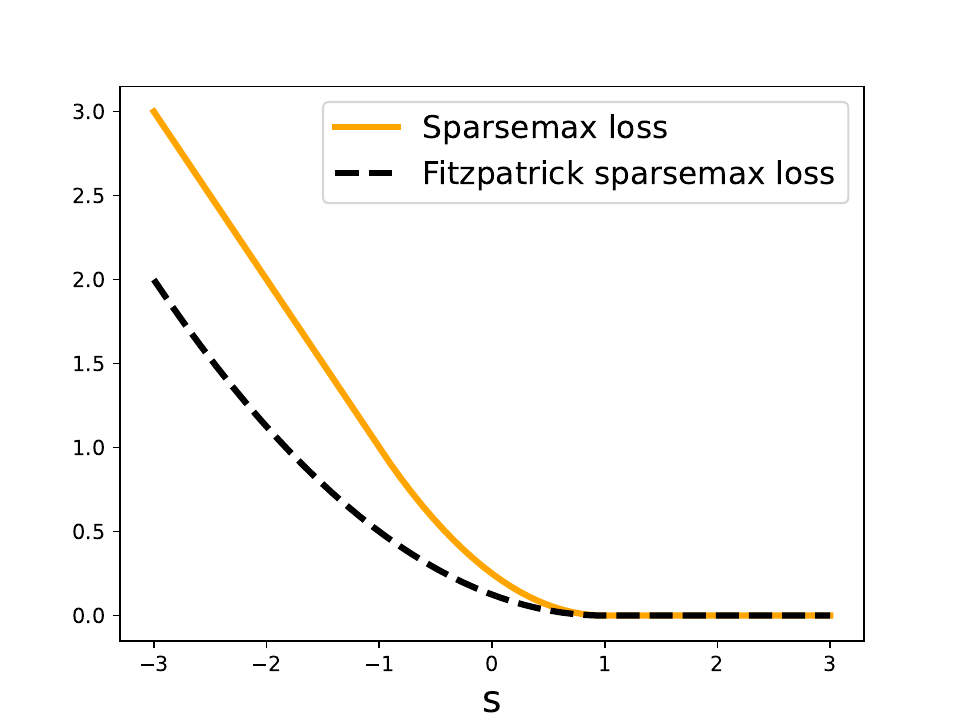}
\caption{
We introduce \textbf{Fitzpatrick losses}, a new family of loss functions
generated by a convex regularization function~$\Omega$, 
that \textbf{lower-bound}
Fenchel-Young losses generated by the same $\Omega$, while maintaining the 
\textbf{same} link function $\widehat{y}_\Omega = \nabla \Omega^*$.  In
particular, we use our framework to instantiate the counterparts of the
\textbf{logistic} and \textbf{sparsemax} losses, two instances of Fenchel-Young
losses, associated with the \textbf{soft argmax} and the \textbf{sparse
argmax}.  In the figures above, we plot $L(y, \theta)$, where $y = e_1$,
$\theta = (s, 0)$ and $L \in \{L_{F[\partial \Omega]}, L_{\Omega \oplus
\Omega^*}\}$, confirming the lower-bound property.
}
\label{fig:lower_bound}
\end{figure}

In this paper, we introduce Fitzpatrick losses, a new family of primal-dual
convex loss functions. Our proposal builds upon the Fitzpatrick function, a
well-known theoretical object in maximal monotone operator theory
\cite{Fitzpatrick:1988,Burachik-Svaiter:2002,Bauschke-Heinz-McLaren-Sendov-Hristo:2005}.
So far, the
Fitzpatrick function had been used as a theoretical tool to represent maximal
monotone operators \cite{ryu2022large} and to construct Bregman-like
primal-primal divergences \cite{Burachik-Martinez-Legaz:2018}, but
it had not been used to construct primal-dual loss functions for machine
learning, as we do.
Crucially, the Fitzpatrick function naturally leads to a
refined Fenchel-Young inequality, making Fitzpatrick losses tighter than
Fenchel-Young losses. Yet, their predictions are produced using the same link
function, suggesting that we can use Fitzpatrick losses as a tighter replacement
for the corresponding Fenchel-Young losses (Figure \ref{fig:lower_bound}). 
We make the following contributions.
\begin{itemize}[topsep=0pt,itemsep=3pt,parsep=3pt,leftmargin=15pt]
\item After reviewing some background,
we introduce Fitzpatrick losses. They can be thought as a tighter version of
Fenchel-Young losses, that use the same link function.

\item We instantiate two new loss functions in this family: the Fitzpatrick
    logistic loss and the Fitzpatrick sparsemax loss. They are the counterparts
    of the logistic and sparsemax losses, two instances of Fenchel-Young losses.
    We therefore obtain two new tighter losses for the soft argmax and the
    sparse argmax, two of the most popular output layers in machine learning.

\item We study in detail the properties of Fitzpatrick losses.
We show that Fitzpatrick losses are equivalent to Fenchel-Young losses with a
modified, target-dependent generating function.

\item We demonstrate the
effectiveness of Fitzpatrick losses for probabilistic classification on $11$
datasets.

\end{itemize}

\section{Background}
\label{sec:background}

\subsection{Convex analysis}

We define $[k] \coloneqq \{1,\dots,k\}$.
We denote the probability simplex by
$\triangle^k \coloneqq \{p \in \RR_+^k \colon \sum_{i=1}^k p_i = 1\}$
and the extended reals by
$\RRb \coloneqq \RR \cup \{+\infty\}$.
We denote the indicator function of a set $\cC$ by
\(\Indicator{\cC}(y) = 0\) if \(y\in \cC\), \(+\infty\) otherwise.
We denote the effective domain of a function 
$\Omega \colon \RR_+^k \to \RRb$ 
by
~\(\dom \Omega := \{y \in \RR^k : \Omega(y) < +\infty \}\).
We denote the Euclidean projection onto a closed convex set~$\cC$ by
$P_\cC(\theta) = \argmin_{y \in \cC} \|y - \theta\|_2^2$.

For a convex function~$\Omega : \RR^k \to \RRb$,
 its \textbf{subdifferential} 
$\partial \Omega$
is defined by
\begin{align}
(y',\theta') \in \partial \Omega
\iff
\theta' \in \partial \Omega(y') 
\iff
\Omega(y) \ge \Omega(y') + \langle y - y', \theta' \rangle ~ \forall y.
\end{align}
When $\Omega$ is differentiable, the subdifferential is a singleton
and we have $\partial \Omega(y') = \{\nabla \Omega(y')\}$.
The \textbf{normal cone} to a set $\cC$ at $y'$ is defined by
\begin{equation}
\theta' \in N_\cC(y') 
\iff
\langle y - y', \theta' \rangle \le 0
\quad \forall y \in \cC
\end{equation}
if $y' \in \cC$
and $N_\cC(y') = \emptyset$ if $y' \not \in \cC$.
The \textbf{Fenchel conjugate}
~\(\Omega^* : \RR^k \to \RRb\) of a
function~\(\Omega: \RR^k \to \RRb\) is defined by
\begin{equation}
\Omega^*(\theta) 
\coloneqq \sup_{y' \in \RR^k} \langle y',\theta \rangle - \Omega(y').
\end{equation} 
From standard convex analysis, when \(\Omega : \RR^k \to \RRb\) is a convex \lsc
(lower semicontinuous) function,
\begin{equation}
\partial \Omega^*(\theta) 
= \argmax_{y' \in \RR^k} \langle y',\theta \rangle - \Omega(y').
\end{equation} 
When the argmax is unique, it is equal to $\nabla \Omega^*(\theta)$.
We define the \textbf{generalized Bregman divergence}~\cite{Kiwiel:1997}
\(D_\Omega : \RR^k \times \RR^k \to \RRb_+\) 
generated by a convex \lsc function~\(\Omega : \RR^k \to \RRb\) by 
\begin{equation}    
D_\Omega(y,y') 
\coloneqq \Omega(y) - \Omega(y') 
- \sup_{\theta' \in \partial \Omega(y')} \langle y - y', \theta' \rangle,
\label{eq:generalized_bregman}
\end{equation}
with the convention \(+\infty + (-\infty) = +\infty\).
When $\Omega$ is differentiable, it recovers the classical \textbf{Bregman
divergence}
\begin{equation}    
D_\Omega(y,y') 
\coloneqq \Omega(y) - \Omega(y') 
- \langle y - y', \nabla \Omega(y') \rangle.
\end{equation}
Both $y$ and $y'$ belong to the \textbf{primal space}.

\subsection{Fenchel-Young losses}

\subsubsection*{Definition and properties}

The \textbf{Fenchel-Young loss}~\(L_{\Omega \oplus \Omega^*} : \RR^k \times
\RR^k \to \RRb \) generated by a convex \lsc function~$\Omega$
\cite{fyl_jmlr} is
\begin{equation}
L_{\Omega \oplus \Omega^*}(y,\theta) 
\coloneqq \Omega \oplus \Omega^*(y,\theta) - \langle y, \theta \rangle
\coloneqq \Omega(y) +  \Omega^*(\theta) - \langle y, \theta \rangle.
\label{de:fenchel_young_loss}
\end{equation}
As its name indicates, it is grounded in the Fenchel-Young inequality
\begin{equation}
\langle y, \theta \rangle \le \Omega(y) + \Omega^*(\theta)    
\quad
\forall y, \theta \in \RR^k.
\end{equation}
The Fenchel-Young loss enjoys many desirable properties,
notably it is \textbf{non-negative} and it is \textbf{convex} in $y$
and $\theta$ separately.
The Fenchel-Young loss can be seen as a \textbf{primal-dual} Bregman divergence
\cite{amari2016information,fyl_jmlr}, 
where $y$ belongs to the primal space and $\theta$ belongs to
the dual space. 

\subsubsection*{Link functions}

To map a dual-space $\theta$ to a primal-space $y$, 
we can use the canonical link function $\partial \Omega^*$, since
\begin{equation}
L_{\Omega \oplus \Omega^*}(y,\theta) = 0
\iff
y \in \partial \Omega^*(\theta).
\end{equation}
In particular when $\Omega$ is strictly convex,
the Fenchel-Young loss is positive, meaning that
it satisfies the identity of indiscernibles
\begin{equation}
L_{\Omega \oplus \Omega^*}(y,\theta) = 0
\iff
y = \nabla \Omega^*(\theta).
\end{equation}
In the remainder of this paper, we will use the notation 
$\widehat{y}_\Omega(\theta)$
to denote the gradient 
$\nabla \Omega^*(\theta)$
or any subgradient in $\partial \Omega^*(\theta)$.
Since $\Omega^*$ is convex, $\widehat{y}_\Omega$ is monotone.
As shown in \cite{fyl_jmlr}, the monotonicity implies that $\theta$
and $\widehat{y}_\Omega(\theta)$ are sorted the same way, i.e.,
$\theta_i > \theta_j \Longrightarrow \widehat{y}_\Omega(\theta)_i \ge
\widehat{y}_\Omega(\theta)_j$.
Link functions also play an important role in the loss gradient, as we have
\begin{equation}
\label{eq:FY_loss_gradient}
\partial_\theta L_{\Omega \oplus \Omega^*}(y, \theta)
= \widehat{y}_\Omega(\theta) - y.
\end{equation}

\subsubsection*{Examples of Fenchel-Young loss instances and their associated
link function}

We give a few examples of Fenchel-Young losses.
With the squared $2$-norm, 
$\Omega(y') = \frac{1}{2} \|y'\|^2_2$,
we obtain the \textbf{squared loss} 
\begin{equation}
L_{\Omega \oplus \Omega^*}(y,\theta)
= L_{\mathrm{squared}}(y, \theta)
\coloneqq \frac{1}{2} \|y - \theta\|^2_2
\end{equation}
and the \textbf{identity link}
\begin{equation}
\widehat{y}_\Omega(\theta) = \theta.
\end{equation}
With the indicator of a convex set $\cC$, 
$\Omega(y') = \Indicator{\cC}(y')$,
we obtain the \textbf{perceptron loss}
\begin{equation}
L_{\Omega \oplus \Omega^*}(y,\theta)
= L_{\mathrm{perceptron}}(y, \theta)
\coloneqq \max_{y' \in \cC} ~ \langle y', \theta \rangle
- \langle y, \theta \rangle.
\end{equation}
and the \textbf{argmax link}
\begin{equation}
\widehat{y}_\Omega(\theta) 
= \argmax_{y \in \cC} ~ \langle y, \theta \rangle. 
\end{equation}
With the squared $2$-norm restricted to some convex set $\cC$, 
$\Omega(y') = \frac{1}{2} \|y'\|^2_2 + \Indicator{\cC}(y')$,
we obtain the \textbf{sparseMAP loss} \cite{niculae2018sparsemap}
\begin{equation}
L_{\Omega \oplus \Omega^*}(y,\theta)
= L_{\mathrm{sparseMAP}}(y, \theta)
\coloneqq \frac{1}{2} \|y - \theta\|^2_2 
- \frac{1}{2} \|P_{\cC}(y) - \theta\|^2_2.
\end{equation}
The link is the \textbf{Euclidean projection} onto $\cC$, 
\begin{equation}
\widehat{y}_\Omega(\theta)
= P_{\cC}(\theta).    
\end{equation}
When the set is $\cC = \triangle^k$,
we obtain the \textbf{sparsemax loss} \cite{martins2016softmax}
and the \textbf{sparsemax link} 
$\widehat{y}_\Omega(\theta) = P_{\triangle^k}(\theta)$,
which is known to produce sparse probability distributions.
With the Shannon negentropy restricted to the probability simplex,
$\Omega(y) \coloneqq \langle y', \log y' \rangle + \Indicator{\triangle^k}(y')$,
we obtain the \textbf{logistic loss}
\begin{equation}
L_{\Omega \oplus \Omega^*}(y,\theta)
= L_{\mathrm{logistic}}(y, \theta)
\coloneqq \log \sum_{i=1}^k \exp(\theta_i) + \langle y, \log y \rangle
- \langle y, \theta \rangle , 
\end{equation}
and the \textbf{soft argmax link} (also know as softmax)
\begin{equation}
\widehat{y}_\Omega(\theta)
= \mathrm{softargmax}(\theta)
\coloneqq \exp(\theta) / \sum_{i=1}^k \exp(\theta_i).    
\end{equation}

\subsection{Maximal monotone operators and the Fitzpatrick function}

An operator $A$ is called \textbf{monotone} if for all $(y, \theta) \in A$
and all $(y',\theta') \in A$, we have
\begin{equation}
\langle y' - y, \theta' - \theta \rangle \ge 0.
\end{equation}
We overload the notation to denote
$A(y) \coloneqq \{\theta \colon (y, \theta) \in A\}$.
A monotone operator $A$ is said to be \textbf{maximal} if there does not
exist $(y,\theta) \not \in A$ such that $A \cup \{(y,\theta)\}$ is still
monotone.
It is well-known that the subdifferential $\partial \Omega$ of a convex function
$\Omega$ is maximal monotone.
For more details on monotone operators,
see \cite{Bauschke-Combettes:2017,ryu2022large}.

A well-known object in monotone operator theory,
the \textbf{Fitzpatrick function}
associated with a monotone operator $A$ 
\cite{Fitzpatrick:1988,Burachik-Svaiter:2002,Bauschke-Heinz-McLaren-Sendov-Hristo:2005},
denoted \(F[A] : \RR^k \times \RR^k \to \RRb\),
is defined by
\begin{equation}
F[A] (y, \theta) \coloneqq
\sup_{(y',\theta') \in A} 
\langle y - y', \theta' \rangle + \langle y', \theta \rangle.
\end{equation}
In particular, with $A=\partial \Omega$, we have
\begin{align}
F[\partial \Omega] (y, \theta)
= \sup_{(y',\theta') \in \partial \Omega}
\langle y - y', \theta' \rangle + \langle y', \theta \rangle
= 
\sup_{y' \in \dom \Omega}
\langle y', \theta \rangle +
\sup_{\theta' \in \partial \Omega(y')}
\langle y - y', \theta' \rangle.
\end{align}
The Fitzpatrick function was studied in depth in
\cite{Bauschke-Heinz-McLaren-Sendov-Hristo:2005}.
In particular, it is jointly convex
and satisfies 
\begin{equation}
\langle y, \theta \rangle 
\le F[\partial \Omega](y, \theta) 
\le \Omega \oplus \Omega^*(y, \theta)
= \Omega(y) + \Omega^*(\theta)
\quad
\forall y, \theta \in \RR^k.
\label{eq:fitzpatrick_ineq}
\end{equation}
From Danskin's theorem, when \(\dom \Omega\) is compact, we also have
\begin{equation}
y^\star_{F[\partial\Omega]}(y, \theta) \coloneqq
\partial_\theta F[\partial\Omega](y, \theta) = \argmax_{y' \in \dom \Omega} ~
\langle y', \theta \rangle +
\sup_{\theta' \in \partial \Omega(y')}
\langle y - y', \theta' \rangle.
\label{eq:fitzpatrick_subgradient}
\end{equation}
The Fitzpatrick function 
$F[\partial \Omega](y,\theta)$ 
and 
$\Omega \oplus \Omega^*(y, \theta) = \Omega(y) + \Omega^*(\theta)$ 
play a similar role but the latter is \textbf{separable} in $y$ and $\theta$,
while the former is \textbf{not}. 
In particular this makes the subdifferential
$\partial_\theta F[\partial \Omega](y, \theta)$ depend on 
both $y$ and $\theta$,
while $\partial_\theta (\Omega \oplus \Omega^*)(y, \theta) = 
\partial \Omega^*(\theta)$ depends only on $\theta$.

The Fitzpatrick function was used in \cite{Burachik-Martinez-Legaz:2018}
to theoretically study primal-primal Bregman-like divergences.
As discussed in more detail in Section \ref{sec:relation_bregman},
using these divergences for machine learning would require us to compose them with
an explicit link function, which would typically break convexity.
In the next section, we introduce new primal-dual losses based on the
Fitzpatrick function.

\section{Fitzpatrick losses}

\subsection{Definition and properties}

Inspired by the inequality in \eqref{eq:fitzpatrick_ineq},
which we can view as a refined Fenchel-Young inequality,
we introduce Fitzpatrick losses,
a new family of loss functions generated by a convex \lsc function $\Omega$.
\begin{definition}{Fitzpatrick loss generated by a convex \lsc function~$\Omega$}
\label{def:fp_loss}

When $y \in \dom \Omega$ and $\theta \in \RR^k$,
we define the Fitzpatrick loss~
\(L_{F[\partial \Omega]} : \RR^k \times \RR^k \to \RRb \) 
generated by a proper convex \lsc function
~\(\Omega : \RR^k \to \RRb\) by
\begin{align}
L_{F[\partial \Omega]}(y,\theta) 
&\coloneqq F[\partial \Omega](y, \theta) - \langle y, \theta \rangle \\
&= \sup_{(y',\theta') \in \partial \Omega}
\langle y - y', \theta' \rangle + \langle y', \theta \rangle 
- \langle y, \theta \rangle\\
&=  \sup_{(y',\theta') \in \partial \Omega}
    \langle y' - y, \theta - \theta' \rangle.
\end{align}
When $y \not \in \dom\Omega$, 
$L_{F[\partial \Omega]}(y,\theta) = +\infty$.
\end{definition}

Fitzpatrick losses enjoy similar properties as Fenchel-Young losses, but they
are \textbf{tighter}.
\begin{proposition}{Properties of Fitzpatrick losses}
\label{prop:properties_fp_losses}
\begin{enumerate}
    \item \textbf{Non-negativity:}
for all $(y, \theta) \in \RR^k$,
$L_{F[\partial \Omega]}(y, \theta) \ge 0$.

    \item \textbf{Same link function:}
        $L_{\Omega \oplus \Omega^*}(y, \theta) =
        L_{F[\partial \Omega]}(y, \theta) = 0 \iff y =
        \widehat{y}_\Omega(\theta)$.

    \item \textbf{Convexity:} $L_{F[\partial \Omega]}(y, \theta)$ is convex in
        $y$ and $\theta$ separately.

    \item \textbf{(Sub-)Gradient:} 
$\partial_\theta L_{F[\partial \Omega]}(y, \theta) 
= y^\star_{F[\partial \Omega]}(y, \theta) - y$ where $y^\star_{F[\partial
\Omega]}(y, \theta)$ is given by~\eqref{eq:fitzpatrick_subgradient}.
\label{item:fp_gradient}

    \item \textbf{Tighter inequality:}
for all $(y, \theta) \in \RR^k$,
$0 
\leq L_{F[\partial\Omega]}(y,\theta) 
\leq L_{\Omega \oplus \Omega^*}(y,\theta)$.
\end{enumerate}
\label{prop:properties}
\end{proposition}
A proof is given in Appendix \ref{proof:prop_properties}.
Because the Fitzpatrick loss and the Fenchel-Young loss generated by the same
$\Omega$ have the same link function $\widehat{y}_\Omega$, 
they share the same minimizers \wrt
$\theta$ for $y$ fixed.  However,
the Fitzpatrick loss is always a \textbf{lower bound} of the 
corresponding Fenchel-Young loss. Moreover, they have different gradients \wrt
$\theta$:
$\partial_\theta L_{\Omega \oplus \Omega^*}(y, \theta)
= \widehat{y}_\Omega(\theta) - y$
vs.
$\partial_\theta L_{F[\partial \Omega]}(y, \theta) 
= y^\star_{F[\partial \Omega]}(y, \theta) - y$.
It is worth noticing that
$y^\star_{F[\partial \Omega]}(y, \theta)$
depends on both $y$ and $\theta$,
contrary to
$\widehat{y}_\Omega(\theta)$.

When $\Omega$ is an unconstrained twice differential function on its domain
(which is for instance the case of the squared $2$-norm or the negentropy),
we next show that Fitzpatrick losses
enjoy a particularly simple expression and become
a squared Mahalanobis-like distance. 
\begin{proposition}{Expressions of $F[\partial \Omega](y, \theta)$
and $L_{F[\partial \Omega]}(y, \theta)$ when $\Omega$ is twice differentiable}
\label{prop:differentiable_Omega}

Suppose $\Omega$ is twice differentiable. Then,
\begin{align}
F[\partial \Omega](y, \theta)
&=
\langle y, \nabla \Omega(y^\star) \rangle 
+ \langle y^\star, \theta \rangle 
- \langle y^\star, \nabla \Omega(y^\star) \rangle \\
L_{F[\partial \Omega]}(y, \theta)
&= \langle y^\star - y, \theta - \nabla \Omega(y^\star) \rangle \\
&= \langle y^\star - y, \nabla^2 \Omega(y^\star)(y^\star-y) \rangle
\end{align}
where $y^\star = y^\star_{F[\partial \Omega]}(y, \theta)$ is the solution
\wrt $y'$ of
\begin{equation}
\nabla^2 \Omega(y')(y'-y) = \theta - \nabla \Omega(y').
\end{equation}
\end{proposition}
A proof is given in \ref{proof:differentiable_Omega}.
When $\Omega$ is constrained (i.e., when it contains an indicator function),
we show in Section \ref{sec:lower_bound} that the above expression becomes a
lower bound.

\subsection{Examples}

We now present the Fitzpatrick loss counterparts of various Fenchel-Young losses.

\paragraph{Squared loss.} 

\begin{proposition}{Squared loss as a Fitzpatrick loss}
\label{prop:squared_loss}

When \(\Omega(y') = \frac{1}{2}\norm{y'}^2_2\), we have  
for all $y \in \RR^k$ and $\theta \in \RR^k$
\begin{equation}
L_{F[\partial \Omega]}(y,\theta) 
= \frac{1}{4} \norm{y-\theta}^2_2
= \frac{1}{2} L_{\mathrm{squared}}(y, \theta).
\end{equation}
\end{proposition}
A proof is given in Appendix \ref{proof:squared_loss}.
Therefore, the Fenchel-Young and Fitzpatrick losses generated by~$\Omega$ 
coincide, up to a factor $\frac{1}{2}$.

\paragraph{Perceptron loss.}

\begin{proposition}{Perceptron loss as a Fitzpatrick loss}
\label{prop:perceptron_loss}

When $\Omega(y') = \Indicator{\cC}(y')$, 
where $\cC$ is a closed convex set,
we have
for all $y \in \cC$ and $\theta \in \RR^k$
\begin{equation}
L_{F[\partial \Omega]}(y,\theta)
= L_{\mathrm{perceptron}}(y, \theta)
= \max_{y' \in \cC} ~ \langle y', \theta \rangle - \langle y, \theta \rangle.
\end{equation}
\end{proposition}
A proof is given in Appendix \ref{proof:perceptron_loss}.
Therefore, the Fenchel-Young and Fitzpatrick losses generated by~$\Omega$ 
exactly coincide in this case.

\paragraph{Fitzpatrick sparseMAP and Fitzpatrick sparsemax losses.} 

As our first example where Fenchel-Young and Fitzpatrick losses substantially
differ, we introduce the \textbf{Fitzpatrick sparseMAP} loss, which is the
Fitzpatrick counterpart of the sparseMAP loss \cite{niculae2018sparsemap}.
\begin{proposition}{Fitzpatrick sparseMAP loss}
\label{prop:fitzpatrick_sparsemap}

When $\Omega(y') = \frac{1}{2} \|y'\|^2_2 + \Indicator{\cC}(y')$,
where $\cC$ is a closed convex set,
we have for all $y \in \cC$ and $\theta \in \RR^k$
\begin{equation}
L_{F[\partial \Omega]}(y,\theta) 
= 2\Omega^*\left((y + \theta) / 2 \right) - \langle y, \theta \rangle \\
= \langle y^\star - y, \theta - y^\star \rangle
\end{equation}
where we used $y^\star$ as a shorthand for
\begin{equation}
y^\star_{F[\partial \Omega]}(y, \theta)
= \nabla \Omega^*((y + \theta) / 2)
= P_{\cC}((y + \theta) / 2).
\end{equation}
\end{proposition}
A proof is given in Appendix \ref{proof:fitzpatrick_sparsemap}.
As a special case, when $\cC = \triangle^k$,
we call the obtained loss the \textbf{Fitzpatrick sparsemax loss}, as it is the
counterpart of the sparsemax loss \cite{martins2016softmax}.
Like the sparseMAP and sparsemax losses, these new losses 
rely on the Euclidean projection as a core building block.
The Euclidean projection onto the probability simplex $\triangle^k$
can be computed exactly in
$O(k)$ expected time and $O(k \log k)$ worst-case time
\cite{Brucker1984,michelot,duchi,Condat2016}.

\paragraph{Fitzpatrick logistic loss.}

We now derive the Fitzpatrick couterpart of the logistic loss.
Before stating the next proposition, 
we recall the definition of the Lambert $W$
function \cite{Corless:1996}. For $z\geq 0$, $W(z)$ is the inverse of the
function $f(w) = w \exp(w)$. That is, $W(z) = f^{-1}(z) = w$.
\begin{proposition}{Fitzpatrick logistic loss}
\label{prop:fitzpatrick_logistic}

    When \(\Omega(y') = \langle y', \log y' \rangle +
    \Indicator{\triangle^k}(y')\),
we have for all $y \in \triangle^k$ and $\theta \in \RR^k$
    \begin{equation}
    \label{eq:formula_fp_loss_multiclass_logistic}
    L_{F[\partial \Omega]}(y,\theta) =  \langle y^\star -  y  ,  \theta -\log y^\star - 1  \rangle 
    \end{equation}
    where we used \(y^\star\) as a shorthand for  \(y^\star_{F[\partial \Omega]}(y,\theta)\) defined by
    \begin{equation}
        y^\star_{F[\partial \Omega]}(y,\theta)_i =  \left\{
            \begin{array}{c}
               \e^{-\lambda^\star} \e^{\theta_i}, \text{ if } y_i =0 , \\
               \frac{y_i}{W(y_i \e^{\lambda^\star - \theta_i})}, \text{ if } y_i >0 .
            \end{array}
        \right.
    \end{equation} 
    
\end{proposition}
A proof and the value of \(\lambda^\star = \lambda^\star(y,\theta) \in \RR\)
are given in Appendix~\ref{proof:formula_fp_loss_logistic}.
To obtain $\lambda^\star(y,\theta)$, we need to solve a one-dimensional root
equation, which can be done using for instance a bisection.

\subsection{Relation with Fenchel-Young losses}

\begin{figure}[t]
\centering
\includegraphics[width=0.75\textwidth]{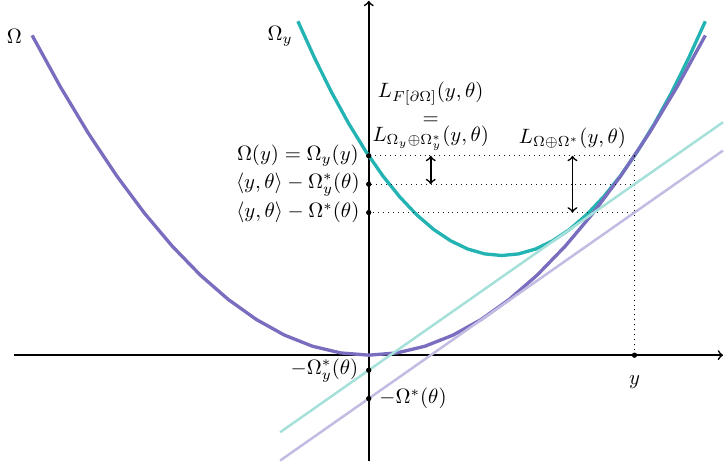}
\caption{
\textbf{Geometric interpretation},
with $\Omega(y') = \frac{1}{2} \|y'\|^2_2$.
The Fenchel-Young loss 
$L_{\Omega \oplus \Omega^*}(y, \theta)$ is the gap (depicted with a
double-headed arrow)  between
$\Omega(y)$ and $\langle y, \theta \rangle - \Omega^*(\theta)$, the value at $y$
of the tangent with slope $\theta$ and intercept $-\Omega^*(\theta)$.
As per Proposition \ref{prop:fitzpatrick_func_generalized_bregman},
the Fitzpatrick loss 
$L_{F[\partial \Omega]}(y, \theta)$ is equal to $L_{\Omega_y \oplus
\Omega_y^*}(y, \theta)$ and is therefore equal to the gap
between
$\Omega_y(y) = \Omega(y)$ and $\langle y, \theta \rangle - \Omega_y^*(\theta)$,
the value at $y$ of the tangent with slope $\theta$ and intercept
$-\Omega_y^*(\theta)$.
Since $\Omega_y(y') = \Omega(y') + D_\Omega(y, y')$, we have
that $\Omega_y(y') \ge \Omega(y')$, with equality when $y = y'$.
We therefore have
$\Omega_y^*(\theta) \le \Omega^*(\theta)$,
implying that the Fitzpatrick loss is a lower bound of the Fenchel-Young
loss.
}
\label{fig:duality_gap}
\end{figure}

On first sight, Fitzpatrick losses and Fenchel-Young losses appear quite
different. 
In the next proposition,
we show that the Fitzpatrick loss generated by $\Omega$ is in fact equal
to the Fenchel-Young loss generated by
the modified, target-dependent function
\begin{equation}
    \Omega_y(y') \coloneqq \Omega(y') + D_\Omega(y, y'),
\end{equation}
where $D_\Omega$ is the generalized Bregman divergence defined 
in \eqref{eq:generalized_bregman}.
In particular, Lemma \ref{lemma:Bregman_constrained_Omega} in the appendix
shows that if $\Omega = \Psi + \iota_\cC$, then 
$\Omega_y(y') 
= \Psi_y(y') + \iota_\cC(y') 
= \Psi(y') + D_\Psi(y, y') + \iota_\cC(y')$.
\begin{proposition}{Characterization of 
$F[\partial \Omega]$,
$L_{F[\partial \Omega]}$
and
$y^\star_{F[\partial \Omega]}$ 
using $\Omega_y$}
\label{prop:fitzpatrick_func_generalized_bregman} 

Let \(\Omega : \RR^k \to \RRb \) be a proper convex \lsc function.
Then, for all $y \in \dom\Omega$ and all $\theta \in \RR^k$,
\begin{align}
F[\partial\Omega](y,\theta) 
&= \Omega_y(y) + \Omega_y^*(\theta) \\
L_{F[\partial\Omega]}(y,\theta) 
&= L_{\Omega_y \oplus \Omega_y^*}(y, \theta)
\\
y^\star_{F[\partial\Omega]}(y, \theta) 
&= \widehat{y}_{\Omega_y}(\theta).
\end{align}
\end{proposition}
This characterization of the Fitzpatrick function $F[\partial \Omega]$ 
is also new to our knowledge.
A proof is given in Appendix \ref{proof:fitzpatrick_func_generalized_bregman}.
Proposition 
\ref{prop:fitzpatrick_func_generalized_bregman} 
is very useful, as it means
that Fitzpatrick losses inherit from all the known properties of Fenchel-Young
losses, analyzed in prior works \cite{fyl_jmlr,Blondel:2022}. 
In particular, Fenchel-Young losses are smooth (i.e., with Lipschitz gradients)
when $\Omega$ is strongly convex. We therefore immediately obtain that
Fitzpatrick losses are smooth if $\Omega$ is strongly convex and $D_\Omega$ is
convex in its second argument, which is the case when $\Omega(y') = \frac{1}{2}
\|y'\|_2^2$ and $\Omega(y') = \langle y', \log y' \rangle$.
Therefore, the Fitzpatrick sparsemax and logistic losses are smooth.
Proposition 
\ref{prop:fitzpatrick_func_generalized_bregman} 
also provides a mean to compute Fitzpatrick losses and their gradient.
Finally, it suggests a very natural geometric interpretation of Fitzpatrick
losses, as presented in Figure \ref{fig:duality_gap}.

\subsection{Relation with generalized Bregman divergences}
\label{sec:relation_bregman}

As we stated before, the generalized Bregman divergence 
$D_\Omega(y, y')$ in \eqref{eq:generalized_bregman} is a primal-primal
divergence, as both $y$ and $y'$ belong to the same primal space.
In contrast, Fenchel-Young losses $L_{\Omega \oplus \Omega^*}(y, \theta)$
are primal-dual, since $y$ belongs to the primal space and $\theta$ belongs to
the dual space. Both can however be related, since
\begin{align*}
D_{\Omega}(y,y') 
&= \inf_{\theta' \in \partial \Omega(y')} L_{\Omega \oplus \Omega^*}(y,\theta') \\
&= \inf_{\theta' \in \partial \Omega(y')} \Omega(y) + \Omega^*(\theta') -
\langle y, \theta' \rangle \\
&= \Omega(y) + \inf_{\theta' \in \partial \Omega(y')} \Omega^*(\theta') -
\langle y, \theta'\rangle \\
&= \Omega(y) - \sup_{\theta' \in \partial \Omega(y')} - \Omega^*(\theta') +
\langle y, \theta'\rangle \\
&= \Omega(y) - \Omega(y') -  \sup_{\theta' \in \partial \Omega(y')}  \langle
y - y', \theta' \rangle,
\end{align*}
where in the last line we used that that \( \Omega^*(\theta') = \langle y',
\theta' \rangle - \Omega(y') \), as \(\theta' \in \partial \Omega(y')\).
This identity suggests that we can create Bregman-like 
primal-primal divergences by replacing $\Omega \oplus \Omega^*$ with 
$F[\partial \Omega]$,
\begin{equation}
\cD_{F[\partial \Omega]}(y,y') \coloneqq
\inf_{\theta' \in \partial \Omega(y')} 
L_{F[\partial \Omega]}(y ,\theta')
=
\inf_{\theta' \in \partial \Omega(y')} 
F[\partial \Omega](y ,\theta') - \langle y, \theta' \rangle.
\end{equation}
This recovers one of the two Bregman-like divergences proposed
in \cite{Burachik-Martinez-Legaz:2018}, the other one replacing the
\(\inf\) above by a \(\sup\).  
As stated in~\cite{Burachik-Martinez-Legaz:2018}, \(F[\partial
\Omega]\) and \(\Omega \oplus \Omega^*\) are \textbf{representations} of
~\(\partial \Omega\). More generally, 
Bregman divergences can be defined for any
representation of the subdifferential~\(\partial \Omega\). 

In order to use a primal-primal divergence as a loss,
we need to explicitly compose it with a link function, 
such as $\widehat{y}_\Omega(\theta) = \nabla \Omega^*(\theta)$.
Unfortunately, 
$D_\Omega(y, \widehat{y}_\Omega(\theta))$
or
$\cD_{F[\partial \Omega]}(y, \widehat{y}_\Omega(\theta))$
are typically \textbf{nonconvex} functions of $\theta$,
while Fenchel-Young and Fitzpatrick losses are always \textbf{convex}.
In addition, differentiating through $\widehat{y}_\Omega(\theta)$ 
typically requires implicit differentiation 
~\cite{krantz2002implicit,blondel_implicit_diff},
while Fenchel-Young and Fitzpatrick losses enjoy easy-to-compute
gradients, thanks to Danskin's theorem.

\subsection{Lower bound on Fitzpatrick losses}
\label{sec:lower_bound}

If 
$\Omega = \Psi + \iota_\cC$, 
where $\Psi$ is a convex Legendre-type function and $\cC \subseteq \dom \Psi$, 
then it was shown in \cite[Proposition 3]{fyl_jmlr} that
Fenchel-Young losses satisfy the lower bound
\begin{equation}
D_\Psi(y, \widehat{y}) \le L_{\Omega \oplus \Omega^*}(y, \theta),
\end{equation}
with equality if $\cC = \dom \Psi$, 
where we used $\widehat{y}$ as a shorthand for
$\widehat{y}_\Omega(\theta)$.
We now show that a similar result holds for Fitzpatrick losses.
\begin{proposition}{Lower bound on Fitzpatrick losses}
\label{prop:lower_bound}

Let $\Omega = \Psi + \iota_\cC$, where $\Psi$ is a convex Legendre-type
function and $\cC \subseteq \dom \Psi$. Then,
\begin{equation}
D_{\Psi_y}(y, y^\star) =
\langle y - y^\star, \nabla^2 \Psi(y^\star) (y - y^\star) \rangle
\le L_{F[\partial \Omega]}(y, \theta),
\end{equation}
with equality if $\dom \Psi = \cC$,
where we used 
$y^\star$ 
as a shorthand for 
$y^\star_{F[\partial \Omega]}(y, \theta)$.
\end{proposition}
A proof is given in Appendix \ref{proof:lower_bound}.
If $\Psi_y$ is $\mu$-strongly convex, we obtain
$\frac{\mu}{2} \|y - y^\star\|^2_2 \le D_{\Psi_y}(y, y^\star)$.

\section{Experiments}
\label{section:experiments}
\paragraph{Experimental setup.} 

We follow exactly the same experimental setup as in 
\cite{fyl_aistats,fyl_jmlr}.
We consider a dataset of \(n\)~pairs \((x_i,y_i)\) of feature vector
\(x_i \in \RR^d\) and label proportions \(y_i \in \triangle^k\), where \(d\)
is the number of features and \(k\) is the number of classes. 
At inference time, given an unknown input vector \(x \in \RR^d\), our goal
is to estimate a vector of label proportions \(\widehat{y} \in \triangle^k\). 
A model is specified by a matrix~\(W \in \RR^{k \times d}\) and a convex \lsc\
function~\(\Omega : \RR^k \to \RRb\).
Predictions are then produced by the generalized linear model
$x \mapsto \widehat{y}_\Omega(Wx)$. 
At training time, we estimate
the matrix \(W \in \RR^{k \times d}\) by minimizing the convex objective
\begin{equation} 
R_{L,\lambda}(W) \coloneqq \sum_{i=1}^n L(y_i, Wx_i) +
\frac{\lambda}{2} \norm{W}^2_2, 
\label{eq:objective}
\end{equation} 
where 
\(L \in \big\{L_{\Omega \oplus \Omega^*}, L_{F[\partial\Omega]} \big\} \).
We focus on the (Fitzpatrick) sparsemax and the (Fitzpatrick) logistic losses.
We optimize \eqref{eq:objective} using the L-BFGS algorithm~\cite{Liu:1989}.
The gradient of the Fenchel-Young loss is given in
\eqref{eq:FY_loss_gradient},
while the gradient of the Fitzpatrick loss is given in Proposition
\ref{prop:properties}, item \ref{item:fp_gradient}.
Experiments were conducted on a Intel Xeon E5-2667 clocked at 3.30GHz with 192
GB of RAM running on Linux. 
Our implementation relies on the SciPy \cite{scipy} and scikit-learn
\cite{sklearn} libraries.

We ran experiments on 11 standard multi-label benchmark datasets\footnote{The
datasets can be downloaded from
\url{http://mulan.sourceforge.net/datasets-mlc.html}
and \url{https://www.csie.ntu.edu.tw/~cjlin/libsvmtools/datasets/}.} (see
Table~\ref{ta:datasets_stats} in Appendix \ref{appendix:datasets_stats} for
statistics on the datasets).
For all datasets, we removed samples with no label, normalized samples to have
zero mean unit variance, and normalized labels to lie in the probability
simplex.  We chose the hyperparameter \(\lambda \in
\{10^{-4},10^{-3},\dots,10^{4}\}\) against the validation set.       
We report test set mean squared error in
Table~\ref{ta:test_accuracy_label_proportion_estimation}.

\paragraph{Results.}

We found that the logistic loss and the Fitzpatrick
logistic loss are comparable on most datasets, with the logistic loss
significantly winning on $2$ datasets and the Fitzpatrick logistic loss
significantly winning on $2$ datasets, out of $11$. Since the Fitzpatrick
logistic loss is slightly more computationally demanding, requiring to solve a
root equation while the logistic loss does not, we believe that that the
logistic loss remains the best choice when we wish to use the softargmax as link
function $\widehat{y}_\Omega$.

Similarly, we found that the sparsemax loss and the Fitzpatrick
sparsemax loss are comparable on most datasets, with the sparsemax loss
significantly winning on only $1$ dataset out $11$ and the Fitzpatrick loss
significantly winning on $2$ datasets out of $11$.  Since the two losses both
use the Euclidean projection onto the simplex $P_{\triangle^k}$ as their link
function $\widehat{y}_\Omega$, we conclude that the Fitzpatrick sparsemax loss
is a serious contender to the sparsemax loss,
especially when predicting sparse label proportions is important.
     
    \begin{table}[t]
      \centering
      \begin{tabular}{r||cc|cc}
        \hline
        Dataset & Sparsemax & Fitzpatrick-sparsemax & Logistic & Fitzpatrick-logistic\\
        \hline
        Birds & 0.531 &  {\bf 0.513} &  0.519 &  0.522 \\      
        Cal500 & {0.035} & {0.035 } & {0.034} & { 0.034} \\
        Delicious & 0.051 & 0.052 & 0.056 &  0.055 \\
        Ecthr A & 0.514 & 0.514 &  0.431 & {\bf 0.423} \\
        Emotions & 0.317 & 0.318 & 0.327 & {\bf0.320} \\
        Flags &  0.186 &  0.188 &  0.184 & 0.187 \\
        Mediamill & {\bf 0.191} & 0.203 & {\bf 0.207} & 0.220 \\
        Scene & 0.363 & {\bf 0.355} & {\bf 0.344} & 0.368 \\
        Tmc & 0.151 & 0.152 & 0.161 &  0.160 \\
        Unfair & 0.149 & 0.148 & 0.157 & 0.158 \\
        Yeast &  0.186 & 0.187 & 0.183 & 0.185 \\
        \hline
      \end{tabular}
    \caption{Test performance comparison between the sparsemax loss, the
        logistic loss and their Fitzpatrick counterparts on the task of label
        proportion estimation, with regularization parameter $\lambda$ tuned 
        against the validation set.
        For each dataset, label proportion errors are measured using the mean
        squared error (MSE). We use bold
        if the error is at least 0.05 lower than its counterpart. 
    }
    \label{ta:test_accuracy_label_proportion_estimation}
    \end{table}

\section{Conclusion}

We proposed to leverage the Fitzpatrick function, a theoretical tool from
monotone operator theory, in order to build a new family of primal-dual convex
loss functions for machine learning.  We showed that Fitzpatrick losses are
lower bounds of Fenchel-Young losses, while maintaining the same link function.
Our paper therefore challenges the idea that there can only be one loss function
associated with a certain link function.  For instance, we created the
Fitzpatrick logistic and sparsemax losses, that are associated with the soft
argmax and sparse argmax links, traditionally associated with the logistic and
sparsemax losses, respectively.  
We believe that even more loss functions with the same
link can be created, which calls for a systematic study of their properties and
respective benefits.

\appendix

\newpage

\section{Datasets statistics}
\label{appendix:datasets_stats} 

\begin{table}[h]
      \centering
      \begin{tabular}{r||ccccccc}
        \hline
        Dataset & Type & Train & Dev & Test & Features & Classes
         & Avg.labels\\
        \hline
        Birds & Audio &  134 & 45 & 172 & 260 & 19 & 2 \\      
        Cal500 & Music & 376 & 126 & 101 & 68 & 174 & 26 \\
        Delicious & Text & 9682 & 3228 &  3181 & 500 & 983 & 19 \\
        Ecthr A & Text & 6683 & 228 & 847 & 92401 & 10 & 1   \\
        Emotions & Music & 293 & 98 & 202 & 72 & 6 & 2\\
        Flags &  Images &  96 & 33 & 65 & 19 & 7 & 3\\
        Mediamill & Video & 22353 & 7451 & 12373 & 120 & 101 & 5\\
        Scene & Images & 908 & 303 & 1196 & 294 & 6 & 1 \\
        Tmc & Text & 16139 & 5380 & 7077 & 48099 & 896 & 6\\
        Unfair & Text & 645 & 215 & 172 & 6290 & 8 & 1 \\
        Yeast &  Micro-array & 1125 & 375 & 917 & 103 & 14 & 4\\
        \hline
      \end{tabular}
    \caption{Datasets statistics}
    \label{ta:datasets_stats}
    \end{table}

\section{Proofs}

\subsection{Lemmas} 

\begin{lemma}{Generalized Bregman divergence for constrained $\Omega$}
\label{lemma:Bregman_constrained_Omega}

Let $\Omega = \Psi + \iota_\cC$, 
where 
$\Psi$ is convex differentiable
and
$\cC \subseteq \dom \Psi$ such that $\mathrm{ri} \cC \cap \mathrm{ri} \dom \Psi \neq \emptyset$,
where $\mathrm{ri} \cC$ is the relative interior of \(\cC\).
Then, for all $y, y' \in \dom \Psi$
\begin{equation}
D_\Omega(y, y') = D_\Psi(y, y') + D_{\iota_\cC}(y, y').
\end{equation}
\end{lemma}
\textbf{Proof.}
As $ \cC, \dom \Psi \subset \RR^k$ and $\mathrm{ri} \cC \cap \mathrm{ri} \dom \Psi \neq \emptyset$, we can apply~\cite[Proposition 6.19]{Bauschke-Combettes:2017} and \cite[Theorem 16.46]{Bauschke-Combettes:2017} to write $\partial \Omega(y') = \partial \Psi(y') + N_\cC(y')$.

Thus, we have
\begin{equation}
\theta' \in \partial \Omega(y')    
\iff
\theta' - \nabla \Psi(y') \in N_\cC(y')
\iff
\delta' \in N_\cC(y'),
\end{equation}
where 
\begin{equation}
\delta' \coloneqq \theta' - \nabla \Psi(y')
\iff
\theta' \coloneqq \delta' + \nabla \Psi(y').
\end{equation}
We then have
\begin{align}    
D_\Omega(y,y') 
&\coloneqq \Omega(y) - \Omega(y') 
- \sup_{\theta' \in \partial \Omega(y')} \langle y - y', \theta' \rangle \\
&= \Omega(y) - \Omega(y') 
- \sup_{\delta' \in N_\cC(y')} \langle y - y', \delta' + \nabla
\Psi(y') \rangle \\
&= \Psi(y) + \iota_\cC(y) - \Psi(y') - \iota_\cC(y')
- \langle y - y', \nabla \Psi(y') \rangle
- \sup_{\delta' \in N_\cC(y')} \langle y - y', \delta' \rangle \\
&= D_\Psi(y, y') + D_{\iota_\cC}(y, y').
\end{align}

\begin{lemma}{Generalized Bregman divergence of indicator function}
\label{lemma:Bregman_indicator}

\begin{equation}
D_{\iota_\cC}(y, y')
= \begin{cases}
    \iota_\cC(y) &\mbox{if } y' \in \cC \\
\infty &\mbox{if } y' \not \in \cC
\end{cases}
= \iota_\cC(y) + \iota_\cC(y').
\end{equation}
\end{lemma}
\textbf{Proof.}
\begin{align}    
D_{\iota_\cC}(y,y') 
&\coloneqq \iota_\cC(y) - \iota_\cC(y') 
- \sup_{\theta' \in N_\cC(y')} \langle y - y', \theta' \rangle.
\end{align}
When $y' \in \cC$ and $y \in \cC$,
\begin{equation}
\sup_{\theta' \in N_\cC(y')}
\langle y - y', \theta' \rangle 
=
\sup_{\substack{\theta' \in \RR^k\\
\langle z - y', \theta' \rangle \le 0\\
\forall z\in\cC}}
\langle y - y', \theta' \rangle 
= 0.
\end{equation}
When $y' \in \cC$ and $y \notin \cC$, $D_{\iota_\cC}(y, y') = +\infty$, as \(+ \infty + (-\infty) = + \infty\) in the definition of the Bregman divergence.
Therefore, when $y' \in \cC$ $D_{\iota_\cC}(y, y') = \iota_\cC(y)$.

When $y' \not \in \cC$, $N_\cC(y') = \emptyset$.
{Again, in the definition of the Bregman divergence,}
\(+ \infty + (-\infty) = + \infty\) 
and we use the convention
\(\sup_{\emptyset} = -\infty\).

\begin{lemma}{Bregman divergence of $\Psi_y$}
\label{lemma:bregman_div_psi_y}

Let $\Psi$ be convex and twice differentiable.
Let $\Psi_y(y') \coloneqq \Psi(y') + D_\Psi(y, y')$. 

Then,
for all $y, y', y'' \in \dom \Psi$,
\begin{align}
D_{\Psi_y}(y', y'')
&= D_\Psi(y, y') - D_\Psi(y, y'') + D(y', y'')
+ \langle y' - y'', \nabla^2 \Psi(y'') (y - y'') \rangle \\
&=
\langle y' - y, \nabla \Psi(y') \rangle -
\langle y' - y, \nabla \Psi(y'') \rangle +
\langle y' - y'', \nabla^2 \Psi(y'') (y - y'') \rangle
\end{align}
and in particular for all $y, y' \in \dom \Psi$
\begin{equation}
D_{\Psi_y}(y, y')
= \langle y - y', \nabla^2 \Psi(y') (y - y') \rangle.
\end{equation}
\end{lemma}

\textbf{Proof.}
For all $y,y' \in \dom \Psi$,
\begin{align}
\Psi_y(y') 
&= \Psi(y') + D_\Psi(y, y') \\
&= \Psi(y') + \Psi(y) - \Psi(y') - \langle y - y', \nabla \Psi(y') \rangle \\
&= \Psi(y) - \langle y - y', \nabla \Psi(y') \rangle.
\end{align}
and therefore
\begin{align}
\nabla \Psi_y(y')
&= -\nabla^2 \Psi(y')y + \nabla \Psi(y') + \nabla^2 \Psi(y')y' \\
&= \nabla^2 \Psi(y')(y' - y) + \nabla \Psi(y').
\end{align}
Therefore, for all $y,y',y'' \in \dom \Psi$,
\begin{align}
D_{\Psi_y}(y', y'')
&= \Psi_y(y') - \Psi_y(y'') - \langle y' - y'', \nabla \Psi_y(y'') \rangle \\
&= \Psi(y') + D_\Psi(y, y') - \Psi(y'') - D_\Psi(y, y'') 
- \langle y' - y'', \nabla^2 \Psi(y'') (y'' - y) \rangle
- \langle y' - y'', \nabla \Psi(y'') \rangle \\
&= D_\Psi(y, y') - D_\Psi(y, y'') + D(y', y'')
+ \langle y' - y'', \nabla^2 \Psi(y'') (y - y'') \rangle \\
&=
\langle y' - y, \nabla \Psi(y') \rangle -
\langle y' - y, \nabla \Psi(y'') \rangle +
\langle y' - y'', \nabla^2 \Psi(y'') (y - y'') \rangle
\end{align}
and in particular,
for all $y,y' \in \dom \Psi$,
\begin{align}
D_{\Psi_y}(y, y')
&= D_\Psi(y, y) - D_\Psi(y, y') + D_\Psi(y, y') 
+ \langle y - y', \nabla^2 \Psi_y(y') (y - y') \rangle \\
&= \langle y - y', \nabla^2 \Psi(y') (y - y') \rangle.
\end{align}

\begin{lemma}{Generalized Bregman divergence of negentropy}
\label{lemma:Bregman_negentropy}

Let $\alpha \in \RR$.
Let $\Psi(y') \coloneqq \sum_{i = 1}^k y_i' \log y_i' - \alpha \sum_{i = 1}^k y_i'$ be defined for \(y'\in \RR^k_+\). Then, for $y,y'\in\RR^k_+$,
\begin{equation}
    D_\Psi(y,y') = \sum_{i=1}^k y_i \log\frac{y_i}{y_i'} - \sum_{i=1}^k (y_i-y_i') 
        + \Indicator{\RR^k_{++}}(y').
\end{equation}
\end{lemma}
\textbf{Proof.} If $y' \in \RR^k_{++}$, $\Psi$ is differentiable at $y'$ and 
    $\nabla \Psi(y')_i = \log y_i + 1 - \alpha$. Thus, \(\partial \Psi = \{\nabla \Psi\}\) and
    \begin{align*}
    D_{\Psi}(y,y') &= \Psi(y) - \Psi(y') - \sup_{\theta' \in \partial \Psi(y')} \langle y-y', \theta'\rangle, \\
     &= \Psi(y) - \Psi(y') - \langle y-y', \nabla \Psi(y')\rangle, \\
     &= \sum_{i=1}^k y_i \log\frac{y_i}{y_i'} - \sum_{i=1}^k (y_i-y_i').
    \end{align*}

    If we prove that \(\partial \Psi(y') = \emptyset\) when there is \(y_i' = 0\),
    we can conclude the proof, as \(\sup_\emptyset = -\infty\) by convention.
    Let us assume that \(y_i'=0\). Suppose that \(\theta' \in \partial \Psi(y')\). Then, by definition 
    of subgradients,
        \begin{equation*}
            \langle y'' - y' , \theta' \rangle + \Psi(y') \leq \Psi(y''), \; \forall y'' \in \RR^k_{++}.
        \end{equation*}
    We choose \(y'' = y' + \varepsilon e_i\), where \(\varepsilon >0\) and \(e_i\) is the \(i\)-th canonical base vector.
    Thus, we obtain
    \begin{align*}
            \varepsilon \theta_i &\leq  \Psi(y' + \varepsilon e_i) - \Psi(y'), \\
            &= \sum_{j=1}^k y'_j \log y'_j + \varepsilon \log \varepsilon - \alpha \sum_{j=1}^k y'_j -\alpha \varepsilon - \big(\sum_{j=1}^k y'_j \log y'_j - \alpha \sum_{j=1}^k y'_j\big),\\
            &= \varepsilon \log \varepsilon - \alpha \varepsilon,
    \end{align*}
    as \(y_i =0\) and \(0 \log 0 = 0\) by convention.
    By noticing that \(\lim_{\varepsilon \to 0^+} \big(\varepsilon \log \varepsilon - \alpha \varepsilon \big)/\varepsilon = -\infty\), we get a contradiction, which concludes the proof.

\begin{lemma}{Value and gradient of $\Psi_y^*$}
\label{lemma:gradient_conjugate_psi_y}

Let $\Psi_y(y') \coloneqq \Psi(y') + D_\Psi(y, y')$, where $\Psi$ is strictly convex and
twice differentiable, $D_\Psi(y, y')$ is convex \wrt $y'$ and $y \in \dom \Psi$. Then, for all $\theta \in \RR^k$,
\begin{align}
\Psi_y^*(\theta)
&= \langle \tilde{y}, \theta \rangle - \Psi(y) + \langle y - \tilde{y}, \nabla
\Psi(\tilde{y}) \rangle \\
\nabla \Psi_y^*(\theta) 
&= \tilde{y}
\end{align}
where $\tilde{y}$ is the solution \wrt $y'$ of
\begin{align}
&\argmax_{y' \in \dom \Psi}
\langle y', \theta \rangle + \langle y - y', \nabla \Psi(y') \rangle \\
&\iff \nabla^2 \Psi(y') (y' - y) = \theta - \nabla \Psi(y').
\end{align}
\end{lemma}

\textbf{Proof.} As \(\Psi \leq \Psi_y\), we have \(\dom \Psi_y \subset \dom \Psi \). Thus, we get
\begin{align}
\Psi_y^*(\theta)
&=
\sup_{y' \in \dom \Psi}
\langle y', \theta \rangle - \Psi_y(y') \\
&=
\sup_{y' \in \dom \Psi}
\langle y', \theta \rangle - (\Psi(y') + \Psi(y) - \Psi(y') - \langle y - y',
\nabla \Psi(y') \rangle) \\
&=
\sup_{y' \in \dom \Psi}
\langle y', \theta \rangle - \Psi(y) + \langle y - y', \nabla \Psi(y') \rangle.
\end{align}
Using Danskin's theorem,
\begin{align}
\nabla \Psi_y^*(\theta)
&=
\argmax_{y' \in \dom \Psi}
\langle y', \theta \rangle - \Psi(y) + \langle y - y', \nabla \Psi(y') \rangle
\\
&=
\argmax_{y' \in \dom \Psi}
\langle y', \theta \rangle + \langle y - y', \nabla \Psi(y') \rangle.
\end{align}
Setting the gradient of the inner function to zero concludes the proof.
\begin{lemma}{Gradient of $\Psi_y^*$, squared norm case}

Let $\Psi(y') \coloneqq \frac{1}{2}\|y'\|^2_2$. Then,
\begin{equation}
\nabla \Psi_y^*(\theta) = \frac{y + \theta}{2}.
\end{equation}
\end{lemma}
\textbf{Proof.} Using Lemma \ref{lemma:gradient_conjugate_psi_y} with $\nabla
\Psi(y') = y'$ and $\nabla^2 \Psi(y') = I$, we obtain
that 
$\nabla \Psi_y^*(\theta)$ is the solution \wrt $y'$ of
$y' - y = \theta - y'$. Rearranging the terms concludes the proof.

\newpage

Before stating the next lemma, we recall the definition of the Lambert $W$
function \cite{Corless:1996}. For $z\geq 0$, $W(z)$ is the inverse of the
function $f(w) = w \exp(w)$. That is, $W(z) = f^{-1}(z) = w$.
\begin{lemma}{Gradient of $\Psi_y^*$, negentropy case}
\label{lemma:gradient_conjugate_psi_y_fp_logistic}

Let $\Psi(y') \coloneqq \sum_{i = 1}^k y_i' \log y_i' - \alpha\sum_{i = 1}^k y_i'$ be defined for \(y'\in \RR^k_+\). Then,
\begin{equation}
\nabla \Psi_y^*(\theta)_i = \left\{
                \begin{array}{cl}
                \e^{\theta_i-2 + \alpha}, &\text{if } y_i = 0 \\
                \frac{y_i}{W(y_i \e^{-(\theta_i-2 + \alpha)})}, 
                &\text{if } y_i > 0.
                \end{array}
            \right.
\end{equation}
\end{lemma}

\textbf{Proof.} 
Using Lemma \ref{lemma:gradient_conjugate_psi_y}, we know that
$\tilde{y}$ is the solution of
$\nabla^2 \Psi(\tilde{y}) (\tilde{y} - y) = \theta - \nabla \Psi(\tilde{y})$.
Using $\nabla \Psi(\tilde{y}) = \log \tilde{y} + 1 - \alpha$ and $\nabla^2
\Psi(\tilde{y}) = 1/\tilde{y}$ (where logarithm and division are performed
element-wise),
we obtain for all $i \in [k]$
\begin{equation}
(\tilde{y}_i - y_i) / \tilde{y}_i = \theta_i - \log \tilde{y}_i - 1 + \alpha
\iff
1 - y_i / \tilde{y}_i = \theta_i - \log \tilde{y}_i - 1 + \alpha.
\end{equation}
When \(y_i =0\), we immediatly have $\tilde y_i = \exp(\theta_i -2 + \alpha)$.
When \(y_i >0\), after rearranging, we obtain
\begin{equation}
\frac{y_i}{\tilde{y}_i} \exp\left(\frac{y_i}{\tilde{y}_i}\right)
= y_i \exp(-(\theta_i - 2 + \alpha))
\iff
\frac{y_i}{\tilde{y}_i}
= W(y_i \exp(-(\theta_i - 2 + \alpha))),
\end{equation}
hence the result.

\begin{lemma}{Gradient of $\Omega_y^*$}
\label{lemma:gradient_of_omega_y}

Let $\Psi$ be a strictly convex function such that
$D_\Psi(y, y')$ is convex \wrt $y'$.
Let 
$\Omega_y(y') 
\coloneqq \Psi_y(y') + \iota_\cC(y') 
= \Psi(y') + D_\Psi(y, y') + \iota_\cC(y')$,
where $\cC \subseteq \dom \Psi$ is closed convex. Then,
\begin{equation}
\nabla \Omega_y^*(\theta) = y^\star
\end{equation}
where $y^\star$ is the solution \wrt $y'$ of
\begin{equation}
\argmax_{y' \in \cC} ~
\langle y', \theta \rangle + \langle y - y', \nabla \Psi(y') \rangle.
\end{equation}
\end{lemma}
\textbf{Proof.} The result again follows from Danskin's theorem.

\begin{lemma}{Dual of simplex-constrained conjugate}
\label{lemma:dual_simplex}

If $\Psi$ is strictly convex 
with $\RR_+^k \subseteq \dom \Psi$, 
 then,
\begin{equation}
(\Psi + \iota_{\triangle^k})^*(\theta) =
\min_{\tau \in \RR}
\tau + (\Psi + \iota_{\RR_+^k})^*(\theta - \tau \ones).
\end{equation}
and
\begin{equation}
\nabla (\Psi + \iota_{\triangle^k})^*(\theta) =
\nabla (\Psi + \iota_{\RR_+^k})^*(\theta - \tau^\star \ones),
\end{equation}
where $\tau^\star$ denotes the optimal dual variable.
\end{lemma}
\textbf{Proof.}
\begin{align}
(\Psi + \iota_{\triangle^k})^*(\theta) 
&= \max_{y' \in \triangle^k} \langle y', \theta \rangle - \Psi(y') \\
&= \max_{y' \in \RR^k_+} \min_{\tau \in \RR}
\langle y', \theta \rangle - \Psi(y') 
- \tau (\langle y', \ones \rangle - 1) \\
&= \min_{\tau \in \RR} \tau + 
\max_{y' \in \RR^k_+} 
\langle y', \theta - \tau \ones \rangle - \Psi(y') \\
&= \min_{\tau \in \RR}
\tau + (\Psi + \iota_{\RR_+^k})^*(\theta - \tau \ones),
\end{align}
where we used that, as the constraints of belonging to the simplex are affine,
they are qualified and we can invert the \(\max\) and the \(\min\).

We use the strict convexity of \(\Psi\) and Danskin's theorem to show that $(\Psi + \iota_{\triangle^k})^*$ is differentiable. Then, we use the converse of Danskin's theorem to conclude.

\newpage
\begin{lemma}{Gradient of $\Omega_y^*$, negentropy, constrained to the simplex}
\label{lemma:gradient_Omega_y_star_negentropy}

Let 
$\Omega = \Psi + \iota_{\triangle^k}$, 
where
$\Psi(y') = \langle y', \log y' \rangle$.
Then,
\begin{equation}
y^\star_i 
= \nabla \Omega_y^*(\theta)_i 
=  \left\{
        \begin{array}{c}
           \e^{-\lambda^\star} \e^{\theta_i}, \text{ if } y_i =0 , \\
           \frac{y_i}{W(y_i \e^{\lambda^\star - \theta_i})}, \text{ if } y_i >0 .
        \end{array}
    \right.
\end{equation}
where $\lambda^\star$ is the solution of
\begin{equation}
    \e^{-\lambda^\star} \sum_{i: y_i =0} \e^{\theta_i} 
    + \sum_{i : y_i>0} \frac{y_i}{W(y_i \e^{-(\theta_i - \lambda^\star)})}
        = 1.
\end{equation}
\end{lemma}
\textbf{Proof.} From Lemma~\ref{lemma:gradient_of_omega_y} and Lemma~\ref{lemma:dual_simplex},
since $\dom \Psi_y = \RR_+^k$, we have
\begin{equation}
y^\star 
= \nabla \Omega_y^*(\theta)
= \nabla \Psi_y^*(\theta - \tau^\star \ones)
\end{equation}
where $\tau^\star$ is the solution of
\begin{equation}
\min_{\tau \in \RR}
\tau + \Psi_y^*(\theta - \tau \ones).
\end{equation}
Setting the gradient of the inner function to zero, we get 
\begin{equation}
\langle \nabla \Psi_y^*(\theta - \tau^\star \ones), \ones \rangle = 1.
\end{equation}
Using Lemma \ref{lemma:gradient_conjugate_psi_y_fp_logistic},
we obtain that $\tau^\star$ satisfies
\begin{equation}
    \e^{-\tau^\star - 2} \sum_{i: y_i =0} \e^{\theta_i} 
    + \sum_{i : y_i>0} \frac{y_i}{W(y_i \e^{-(\theta_i - \tau^\star - 2)})}
        = 1.
\end{equation}
Using the change of variable $\tau^\star = \lambda^\star + 2$ concludes the
proof.

\subsection{Proof of Proposition \ref{prop:properties} (Properties of
Fitzpatrick losses)}
\label{proof:prop_properties}

Apart from differentiability, the proofs follow from the study of Fitzpatrick
functions found in 
\cite{Fitzpatrick:1988,Bauschke-Heinz-McLaren-Sendov-Hristo:2005,ryu2022large}. 
We include the proofs for completeness.

\paragraph{Link function and non-negativity.}

We recall that
\begin{align}
L_{F[\partial \Omega]}(y, \theta)
&=  \sup_{(y',\theta') \in \partial \Omega}
    \langle y' - y, \theta - \theta' \rangle \\
&= -\inf_{(y',\theta') \in \partial \Omega} 
\langle y' - y, \theta' - \theta \rangle.
\end{align}
From the monotonicity of $\partial \Omega$, we have
that if 
$(y,\theta) \in \partial \Omega$ 
and 
$(y',\theta') \in \partial \Omega$, 
then
$\langle y' - y, \theta' - \theta \rangle \ge 0$.
Therefore, for all $(y,\theta) \in \partial \Omega$,
\begin{equation}
\inf_{(y',\theta') \in \partial \Omega} 
\langle y' - y, \theta' - \theta \rangle = 0,
\end{equation}
with the infimum being attained at $(y',\theta')=(y,\theta)$.
This proves the link function.

From the maximality of $\partial \Omega$, 
if $(y,\theta) \not\in \partial \Omega$,
there exists $(y',\theta') \in \partial \Omega$ such that
$\langle y' - y, \theta' - \theta \rangle < 0$.
Therefore, for all $(y,\theta) \not \in \partial \Omega$,
\begin{equation}
\inf_{(y',\theta') \in \partial \Omega} 
\langle y' - y, \theta' - \theta \rangle < 0.
\end{equation}
This proves the non-negativity.

\paragraph{Convexity.}

We recall that
\begin{equation}
L_{F[\partial \Omega]}(y, \theta) 
= F[\partial \Omega] (y, \theta) - \langle y, \theta \rangle
\end{equation}
where
\begin{equation}
F[\partial \Omega](y, \theta)
= \sup_{(y',\theta') \in \partial \Omega}
\langle y - y', \theta' \rangle + \langle y', \theta \rangle 
= \sup_{(y',\theta') \in \partial \Omega}
\langle y', \theta \rangle + \langle y, \theta' \rangle 
- \langle y' , \theta'\rangle.
\end{equation}
The function 
$(y,\theta) \mapsto \langle y', \theta \rangle + \langle y, \theta' \rangle 
- \langle y' , \theta'\rangle$ is jointly convex in $(y,\theta)$ for all
$(y',\theta')$. Since the supremum preserves convexity, 
$F[\partial \Omega] (y, \theta)$ is jointly convex in $(y, \theta)$.
The function $\langle y, \theta \rangle$ is separately
convex / concave in $y$ and $\theta$ but not jointly convex / concave 
in $(y, \theta)$.  Therefore,
$L_{F[\partial \Omega]}(y, \theta)$ is separately convex in $y$ and $\theta$.

\paragraph{Differentiability.}

Since $\Omega(y')$ is strictly convex and $y' \mapsto D_\Omega(y, y')$ is convex,
$\Omega_y(y') = \Omega(y') + D_\Omega(y, y')$ is strictly convex in $y'$.
From the duality between strict convexity and differentiability,
$\Omega_y^*(\theta)$ is differentiable in $\theta$.

\paragraph{Tighter inequality.}

Using
\begin{equation}
\partial \Omega = \{(y',\theta') \colon 
\Omega(y) \ge \Omega(y') + \langle y - y', \theta' \rangle ~ \forall y\}
\end{equation}
and
\begin{equation}
\Omega^*(\theta) 
= \sup_{y' \in \RR^k} \langle y', \theta \rangle - \Omega(y'),
\end{equation}
we get for any $(y',\theta') \in \partial \Omega$,
\begin{align}
\langle y - y', \theta' \rangle + \langle y', \theta \rangle
&\le \Omega(y) - \Omega(y') + \langle y', \theta \rangle \\
&\le \Omega(y) + \Omega^*(\theta).
\end{align}
Therefore
\begin{equation}
F[\partial \Omega](y, \theta)
= \sup_{(y',\theta') \in \partial \Omega}
\langle y - y', \theta' \rangle + \langle y', \theta \rangle 
\le \Omega(y) + \Omega^*(\theta).
\end{equation}

\subsection{Proof of Proposition~\ref{prop:differentiable_Omega} (Expression of Fitzpatrick loss when \(\Omega\) is twice differentiable) }
\label{proof:differentiable_Omega}

We recall that
\begin{equation}
F[\partial \Omega](y, \theta)
= \sup_{(y', \theta') \in \partial \Omega} 
\langle y, \theta' \rangle 
+ \langle y', \theta \rangle 
- \langle y', \theta' \rangle
= \sup_{y' \in \dom \Omega} 
\langle y', \theta \rangle +
\sup_{\theta' \in \partial \Omega(y')}
\langle y, \theta' \rangle 
- \langle y', \theta' \rangle.
\end{equation}
Since $\Omega$ is differentiable, 
we have $\partial \Omega(y') = \{\nabla \Omega(y')\}$
and therefore $\theta' = \nabla \Omega(y')$, which gives
\begin{equation}
F[\partial \Omega](y, \theta)
= \sup_{y' \in \RR^k} 
\langle y, \nabla \Omega(y') \rangle 
+ \langle y', \theta \rangle 
- \langle y', \nabla \Omega(y') \rangle.
\end{equation}
Setting the gradient of the inner function \wrt $y'$ to zero,
we get
\begin{equation}
\nabla^2 \Omega(y')y + \theta - \nabla \Omega(y') 
- \nabla^2 \Omega(y')y' = 0.
\end{equation}
Using the $y' = y^\star$ and $\theta'= \nabla \Omega(y')$ in \ref{def:fp_loss},
we then obtain
\begin{align}
L_{F[\partial \Omega]}(y, \theta)
&= \langle y' - y, \theta - \theta' \rangle \\
&= \langle y' - y, \theta - \nabla \Omega(y') \rangle \\
&= \langle y' - y, \nabla^2 \Omega(y')(y'-y) \rangle.
\end{align}

\subsection{Proof of Proposition \ref{prop:squared_loss} (squared loss)}
\label{proof:squared_loss}

Using Proposition \ref{prop:differentiable_Omega} with
$\nabla \Omega(y') = y'$ and $\nabla^2 \Omega(y') = I$,
we obtain
\begin{equation}
y + \theta - 2y' = 0 
\iff
y' = \frac{y + \theta}{2}.
\end{equation}
We therefore obtain
\begin{align}
L_{F[\partial \Omega]}(y, \theta)
&= \left\langle \frac{y+\theta}{2} - y, 
\theta - \frac{y+\theta}{2} \right\rangle \\
&= \left\langle \frac{\theta - y}{2}, 
\frac{\theta - y}{2} \right\rangle \\
&= \frac{1}{4} \|y - \theta\|^2_2.
\end{align}

\subsection{Proof of Proposition \ref{prop:perceptron_loss} (perceptron loss)}
\label{proof:perceptron_loss}

A proof of the Fitzpatrick function for this case was given in 
\cite[Example 3.1]{Bauschke-Heinz-McLaren-Sendov-Hristo:2005}.
We include a proof for completeness.
Since $\Omega = \iota_\cC$,
we have $\partial \Omega = N_\cC$ and $\dom \Omega = \cC$. 
Therefore, for all $y \in \cC$ and $\theta \in \RR^k$,
\begin{align}
F[\partial \Omega] (y, \theta)
&= 
\sup_{y' \in \dom\Omega}
\langle y', \theta \rangle +
\sup_{\theta' \in \partial \Omega(y')}
\langle y - y', \theta' \rangle \\
&= 
\sup_{y' \in \cC}
\langle y', \theta \rangle +
\sup_{\theta' \in N_\cC(y')}
\langle y - y', \theta' \rangle \\
&=
\sup_{y' \in \cC}
\langle y', \theta \rangle -
\left(
\iota_\cC(y) - \iota_\cC(y') -
\sup_{\theta' \in N_\cC(y')}
\langle y - y', \theta' \rangle
\right) \\
&= 
\sup_{y' \in \cC}
\langle y', \theta \rangle -
D_{\iota_\cC}(y, y') \\
&= 
\sup_{y' \in \cC}
\langle y', \theta \rangle,
\end{align}
where 
in the third line we used that $\iota_\cC(y) = \iota_\cC(y') = 0$
and
where in the last line
we used Lemma \ref{lemma:Bregman_indicator}.
Therefore, for all $y \in \RR^k$ and $\theta \in \RR^k$,
\begin{align}
F[\partial \Omega] (y, \theta)
= \sup_{y' \in \cC} \langle y', \theta \rangle + \iota_\cC(y)
= \iota_\cC(y) + \iota_\cC^*(\theta).
\end{align}

\subsection{Proof of Proposition \ref{prop:fitzpatrick_sparsemap}
(Fitzpatrick sparseMAP loss)}
\label{proof:fitzpatrick_sparsemap}

A proof of the Fitzpatrick function for this case was given in 
\cite[Example 3.13]{Bauschke-Heinz-McLaren-Sendov-Hristo:2005}.
We provide an alternative proof.

From Proposition \ref{prop:fitzpatrick_func_generalized_bregman},
we know that
\begin{equation}
F[\partial \Omega](y, \theta)
= \Omega_y(y) + \Omega_y^*(\theta)
= \Omega(y) + \Omega_y^*(\theta),
\end{equation}
where
\begin{align}
\Omega_y(y') 
&= \frac{1}{2} \|y'\|^2_2 + \frac{1}{2} \|y - y'\|^2_2 + \iota_\cC(y') \\
&= \|y'\|^2_2 + \frac{1}{2} \|y\|^2_2 - \langle y, y' \rangle + \iota_\cC(y') \\
&= 2 \Omega(y') + \Omega(y) - \langle y, y' \rangle.
\end{align}
Using conjugate calculus, we obtain
\begin{equation}
\Omega_y^*(\theta) = 2 \Omega^*\left(\frac{y + \theta}{2}\right) - \Omega(y).
\end{equation}
Therefore,
\begin{equation}
F[\partial \Omega](y, \theta)
= 2 \Omega^*\left(\frac{y + \theta}{2}\right).
\end{equation}
From Proposition \ref{prop:fitzpatrick_func_generalized_bregman},
the supremum \wrt $y'$ is achieved
at $y^\star = \nabla \Omega^*((y + \theta) / 2) = P_\cC((y + \theta) / 2)$.
We therefore obtain
\begin{equation}
L_{F[\partial \Omega]}(y,\theta) 
= \langle y^\star - y, \theta - y^\star \rangle.
\end{equation}

\subsection{Proof of Proposition \ref{prop:fitzpatrick_logistic} (Fitzpatrick logistic loss)}
\label{proof:formula_fp_loss_logistic}
    
    \paragraph{Differentiability \wrt $\theta$ and formula of gradient.}
    According to Proposition~\ref{prop:fitzpatrick_func_generalized_bregman}, we have 
    \begin{equation}
        L_{F[\partial \Omega]}(y,\theta) = \Omega_y(y) + 
        \Omega_y^*(\theta) - \langle y, \theta \rangle.
    \end{equation}
    Thus the differentiability \wrt \(\theta\) of $L_{F[\partial \Omega]}(y,\theta)$ follows from the differentiability of $\Omega_y^*(\theta)$.
    Lemma~\ref{lemma:gradient_Omega_y_star_negentropy} yields 
    the differentiability of $\Omega_y^*(\theta)$ and a formula for its
    gradient~\(y^\star_{F[\partial \Omega]}(y,\theta) \coloneqq~\nabla \Omega_y^*(\theta) \).

    \begin{equation}
        y^\star_{F[\partial \Omega]}(y,\theta)_i =  \left\{
            \begin{array}{c}
               \e^{-\lambda^\star} \e^{\theta_i}, \text{ if } y_i =0 , \\
               \frac{y_i}{W(y_i \e^{\lambda^\star - \theta_i})}, \text{ if } y_i >0 .
            \end{array}
        \right.
    \end{equation}
    It follows that \(\nabla_\theta L_{F[\partial \Omega]}(y,\theta) = y^\star_{F[\partial \Omega]}(y,\theta) - y\).

    \paragraph{Formula of the Fitzpatrick logistic loss.}
    We use $y^\star$ as a shorthand for $y^\star_{F[\partial \Omega]}(y,\theta)$. As we know that
     \(\Omega_y^*(\theta) = \langle y^\star, \theta \rangle 
        - \Omega_y(y^\star),
     \)
     we use again Proposition~\ref{prop:fitzpatrick_func_generalized_bregman} to get
     \begin{align*}
        L_{F[\partial \Omega]}(y,\theta) &= 
            \Omega_y(y) + 
        \langle y^\star, \theta \rangle 
        - \Omega_y(y^\star) - \langle y, \theta \rangle \\
        &= \Omega(y) - (\Omega(y^\star) + D_{\Omega}(y,y^\star))  + \langle y^\star - y, \theta \rangle
     \end{align*}
     as \(\Omega_y(y') = \Omega(y) + D_{\Omega}(y,y')\) and in particular
     $\Omega_y(y) = \Omega(y)$. Furthermore, as $y^\star \in \triangle^k \cap \RR^k_{++}$, \(\Omega\) is differentiable at $y^\star$ and 
     \( D_{\Omega}(y,y^\star) = \Omega(y) - \Omega(y^\star) - \langle y - y^\star, \nabla \Omega(y^\star) \rangle \), where
     \(\nabla \Omega(y^\star) = \log y^\star + 1\).
    Thus
    \begin{align*}
        L_{F[\partial \Omega]}(y,\theta) &= 
             \langle y - y^\star, \nabla \Omega(y^\star) \rangle  + \langle y^\star - y, \theta \rangle\\
        &=  \langle y^\star - y, \theta - \log y^\star - 1\rangle.
     \end{align*}

    \paragraph{Bisection formula for \(\lambda^\star\) and bounds.}

    We also get from Lemma~\ref{lemma:gradient_Omega_y_star_negentropy}
    a bisection formula for \(\lambda^\star\), which is a shorthand for 
    \(\lambda^\star_{F[\partial \Omega]}(y,\theta)\).
    \begin{equation}
    \e^{-\lambda^\star} \sum_{i: y_i =0} \e^{\theta_i} 
    + \sum_{i : y_i>0} \frac{y_i}{W(y_i \e^{-(\theta_i - \lambda^\star)})}
        = 1.
    \end{equation}

    We focus here on a lower bound and an upper bound for \(\lambda^\star \in \RR\).
    Let us prove that 
    \begin{equation}
         \log\sum_{i=1}^k \e^{\theta_i} 
         \leq \lambda^\star  \leq \log 2 +
         \max \Big\{
             \log\sum_{i: y_i = 0}^k \e^{\theta_i}, \log \ell_0(y) +
            \max_{i: y_i >0} \theta_i + 2 \ell_0(y) y_i 
         \Big\},
        \end{equation}        
        where \(\ell_0(y)= \mathrm{Card}(j : y_j \neq 0)\).

        \underline{For the lower bound}, we use the concavity of the Lambert function~\(W\), which implies
        \(\frac{1}{W(y_i \e^{\lambda^\star - \theta_i})} \geq \frac{1}{y_i \e^{\lambda^\star - \theta_i}}  \). Thus, 
        \begin{equation}
           1 \geq \e^{-\lambda^\star} \sum_{i: y_i =0} \e^{\theta_i} 
            + \sum_{i : y_i>0} \frac{y_i}{y_i \e^{\lambda^\star - \theta_i}},
        \end{equation}
        which in turn implies
        \begin{equation}
           \e^{\lambda^\star} \geq  \sum_{i: y_i =0} \e^{\theta_i} 
            + \sum_{i : y_i>0} \e^{\theta_i}
        \end{equation}
        and yields the lower bound.

        \underline{For the upper bound}, the function \(g(\lambda) = \e^{-\lambda} \sum_{i: y_i=1} \e^{\theta_i}
            + \sum_{i : y_i>0} \frac{y_i}{W(y_i \e^{\lambda^ - \theta_i})}\) is continuous and decreasing (as it is a positive combination of decreasing functions) and \(g(-\infty) = +\infty\).
            Thus if we find a \(\lambda\) such that \(g(\lambda) < 1\), we know that \(\lambda^\star \leq \lambda\). 

            We deal with each term of \(g(\lambda)\) separately. If \(\lambda \in \RR\) satisfies
            \begin{align*}
                \e^{-\lambda} \sum_{i: y_i =0} \e^{\theta_i} &\leq \frac{1}{2} \\
                \max_{i:y_i >0} \frac{y_i}{W(y_i \e^{\lambda - \theta_i})} &\leq \frac{1}{2 \ell_0(y)},
            \end{align*}
            then 
            \begin{equation}
                g(\lambda) =  \underbrace{\e^{-\lambda} \sum_{i: y_i =0} \e^{\theta_i}}_{\leq 1/2} 
                +  \sum_{i : y_i>0} \underbrace{\frac{y_i}{W(y_i \e^{\lambda^ - \theta_i})}}_{\leq 1/\big(2 \ell_0(y)\big)}   
                \leq 1.        
            \end{equation}
            Thus, all \(\lambda\) satisfying the following inequalties are upper bounds of \(\lambda^\star\)
            \begin{align*}
                2 \sum_{i: y_i =0} \e^{\theta_i} &\leq \e^{\lambda}  \\
                2 \ell_0(y)y_i &\leq W(y_i \e^{\lambda - \theta_i}), \forall i : y_i >0.
            \end{align*}
            As \(W\) is monotone and \(W^{-1}(t) = t \e^t\), we get
            \begin{align*}
                \log 2 + \log \sum_{i: y_i =0} \e^{\theta_i} &\leq \lambda \\
                2 \ell_0(y) \e^{2 \ell_0(y) y_i} &\leq \e^{\lambda^\star - \theta_i}, \forall i : y_i >0.
            \end{align*}

            Thus taking \(
             \lambda = \max \Big\{ \log 2 + 
             \log\sum_{i: y_i = 0}^k \e^{\theta_i}, 
            \max_{i: y_i >0} \log 2 + \log \ell_0(y) + \theta_i + 2 \ell_0(y) y_i
            \Big\}
            \)
            yields an upper bound of \(\lambda^\star\).
         
\subsection{Proof of Proposition \ref{prop:fitzpatrick_func_generalized_bregman} 
(characterization of $F[\partial \Omega]$ using $D_\Omega$)}
\label{proof:fitzpatrick_func_generalized_bregman}

Let \( (y,\theta) \in \dom \Omega \times \RR^k\). 
We have 
\begin{align*}
F[\partial \Omega] (y, \theta)
&=
\sup_{(y',\theta') \in \partial \Omega}
\langle y - y', \theta' \rangle + \langle y', \theta \rangle \\
&= \sup_{y' \in \dom\Omega} \left\{
\langle y',\theta \rangle
+ \sup_{\theta' \in \partial \Omega(y')}
\langle y-y',\theta' \rangle
\right\} \\
&= \sup_{y' \in \dom \Omega}
\left\{
\langle y',\theta \rangle - \Omega(y') + \Omega(y')
+ \sup_{\theta' \in \partial \Omega(y')} \langle y-y',\theta' \rangle
\right\} \\
&= \Omega(y) 
+ \sup_{y' \in \dom \Omega} \langle y',\theta \rangle
- \left(\Omega(y') + \Omega(y) - \Omega(y')
- \sup_{\theta' \in \partial \Omega(y')} \langle y-y',\theta' \rangle
\right) \\
&= \Omega(y) 
+ \sup_{y' \in \dom \Omega} \langle y',\theta \rangle
- \left(\Omega(y') + D_\Omega(y, y') \right) \\
&= \Omega(y) + \left(\Omega + D_{\Omega}(y,\cdot)\right)^*(\theta) \\
&= \Omega_y(y) + \Omega_y^*(\theta).
\end{align*}
The supremum above is achieved at 
$y' \in \partial \Omega_y^*(\theta) 
= y^\star_{F[\partial \Omega]}(y, \theta)$.

When $\Omega = \Psi + \iota_\cC$, where $\cC \subseteq \dom \Psi$, 
using Lemma \ref{lemma:Bregman_constrained_Omega} and
\ref{lemma:Bregman_indicator}, we have for all $y \in \cC$
\begin{equation}
\Omega_y(y') = \Psi(y') + D_\Psi(y, y') + \iota_{\cC}(y').
\end{equation}

\subsection{Proof of Proposition \ref{prop:lower_bound}
(lower bound)}
\label{proof:lower_bound}

It was shown in \cite[Proposition 3]{fyl_jmlr} that if $f = g +
\iota_\cC$, where $g$ is Legendre type
with $\cC \subseteq \dom \Psi$, 
then for all $y \in \cC$ and $\theta \in \RR^k$,
\begin{equation}
0 
\le D_g(y, \nabla f^*(\theta)) 
\le L_{f \oplus f^*}(y, \theta),
\end{equation}
with equality if $\cC = \dom g$.
Using $g = \Psi_y$, $f = \Omega_y = \Psi_y + \iota_\cC$,
$y^\star = \nabla \Omega_y^*(\theta) = y^\star_{F[\partial \Omega]}(y, \theta)$,
and Lemma \ref{lemma:bregman_div_psi_y},
we therefore obtain
\begin{equation}
D_{\Psi_y}(y, y^\star) =
\langle y - y^\star, \nabla^2 \Psi(y^\star) (y - y^\star) \rangle
\le L_{\Omega_y \oplus \Omega_y^*}(y, \theta)
= L_{F[\partial \Omega]}(y, \theta).
\end{equation}
If $\Psi_y$ is $\mu$-strongly convex and $D_\Psi$ is convex in its second
argument, then $\Psi_y$ is $\mu$-strongly convex as well.  
Therefore, we also have
\begin{equation}
\frac{\mu}{2} \|y - y^\star\|^2_2 \le D_{\Psi_y}(y, y^\star).
\end{equation}

\end{document}